\documentclass[sigconf]{acmart}
\AtBeginDocument{%
  }

\copyrightyear{2025}
\acmYear{2025}
\setcopyright{acmlicensed}
\acmConference[Preprint] {Proceedings of the 33rd ACM International Conference on Multimedia}{2025}{}
\acmBooktitle{Proceedings of the 33rd ACM International Conference on Multimedia (MM '25), October 27--31, 2025, Dublin, Ireland}
\acmISBN{979-8-4007-2035-2/2025/10}
\acmDOI{10.1145/XXXXXX.XXXXXX}
\renewcommand\footnotetextcopyrightpermission[1]{}

\usepackage{multirow}
\usepackage{multicol}
\usepackage{adjustbox}

\usepackage{pifont}
\usepackage{bbding}

\usepackage{balance}

\usepackage{booktabs}
\usepackage{array}
\newcolumntype{C}[1]{>{\centering\arraybackslash}m{#1}}

\usepackage{algorithm}
\usepackage{algorithmic}

\begin{document}

\title{Learning Long-Range Action Representation by Two-Stream Mamba Pyramid Network for Figure Skating Assessment}

\author{Fengshun Wang}
\affiliation{%
  \institution{Capital University of \\Physical Education and Sports}
  \city{Beijing}
  \country{China}
}
\email{wangfengshun2023@cupes.edu.cn}

\author{Qiurui Wang}
\authornote{ Indicates Corresponding Author.}
\affiliation{%
  \institution{Capital University of \\Physical Education and Sports}
  \city{Beijing}
  \country{China}
}
\email{wangqiurui@cupes.edu.cn}

\author{Peilin Zhao}
\affiliation{%
  \institution{Shanghai Jiao Tong University}
  \city{Shanghai}
  \country{China}
}
\email{peilinzhao@hotmail.com}


\begin{abstract}
Technical Element Score (TES) and Program Component Score (PCS) evaluations in figure skating demand precise assessment of athletic actions and artistic interpretation, respectively. Existing methods face three major challenges. Firstly, video and audio cues are regarded as common features for both TES and PCS predictions in previous works without considering the prior evaluation criterion of figure skating. Secondly, action elements in competitions are separated in time, TES should be derived from each element's score, but existing methods try to give an overall TES prediction without evaluating each action element. Thirdly, lengthy competition videos make it difficult and inefficient to handle long-range contexts. To address these challenges, we propose a two-stream Mamba pyramid network that aligns with actual judging criteria to predict TES and PCS by separating visual-feature based TES evaluation stream from audio-visual-feature based PCS evaluation stream. In the PCS evaluation stream, we introduce a multi-level fusion mechanism to guarantee that video-based features remain unaffected when assessing TES, and enhance PCS estimation by fusing visual and auditory cues across each contextual level of the pyramid.
In the TES evaluation stream, the multi-scale Mamba pyramid and TES head we proposed effectively address the challenges of localizing and evaluating action elements with various temporal scales and give score predictions. With Mamba’s superior ability to capture long-range dependencies and its linear computational complexity, our method is ideal for handling lengthy figure skating videos. Comprehensive experimentation demonstrates that our framework attains state-of-the-art performance on the FineFS benchmark. Furthermore, it yields competitive outcomes on two additional datasets without further training. Our source code is available at \href{https://github.com/ycwfs/Figure-Skating-Action-Quality-Assessment}{https://github.com/ycwfs/Figure-Skating-Action-Quality-Assessment}.

\end{abstract}




\maketitle

\section{Introduction}
The combination of artistic expression and complicated movements in figure skating presents unique challenges for action quality assessment. As one of the most visually complex winter sports, figure skating demands judging systems that can accurately evaluate technical skill and artistic interpretation. Recent advances in computer vision have introduced new possibilities for objective scoring. However, existing methods often struggle to accommodate the sport's dual-assessment approach, where TES and PCS are based on fundamentally different evaluation criteria.

The previous integration of audio-visual modalities in figure skating assessment presents a fundamental mismatch with competition judging protocols. According to ISU judging standards\footnote{https://www.isu.org/}, TES should be evaluated purely through visual contexts and PCS contains musical interpretation. However, existing multi-modal approaches indiscriminately fuse visual and audio features to both TES and PCS assessments.

In recent years, numerous deep learning-based models have been developed for TES and PCS assessments of figure skating. Early approaches focus on extracting spatial-temporal features from raw video data \cite{parmarLearningScoreOlympic2017,xuLearningScoreFigure2020}. 
In addition, pose estimation techniques have played a crucial role in the refinement of movement analysis \cite{liSkeletonBasedAction2021}. However, accurately evaluating TES requires a more fine-grained analysis of action elements. Therefore, it is essential to locate and evaluate each action element throughout the video. 
Some methods, like \cite{jiLocalizationassistedUncertaintyScore2023}, try to use an uncertainty score disentanglement approach combined with subaction localization. 
However, their localization method is not an integrated trainable component of the framework and only predicts the total score for all action elements.

Furthermore, the duration of some figure skating competition routines stays more than 2 minutes, demanding an exceptionally long-range temporal modeling capability that conventional CNN and RNN models can hardly provide. The transformer-based method like \cite{duLearningSemanticsGuidedRepresentations2023}, can handle this challenge with its context modeling ability with large computational complexity.

All the problems mentioned above make us try to propose a framework for TES and PCS assessments of figure skating in one network.
To address the issue that TES assessment has no relevance with audio cues while PCS should consider both video and audio contexts, we propose a two-stream network for TES and PCS predictions where the TES evaluation stream utilizes visual features and PCS evaluation stream exploits visual-audio features, which aligns with actual judging criteria. By leveraging the inherent relationship between auditory and visual modalities, our approach enhances the robustness and accuracy of PCS estimation while ensuring that the evaluation of TES remains unaffected by audio-based features. 
In order to precisely predict TES, a multi-scale Mamba pyramid is adopted to accurately localize, classify, and score action elements with various durations in figure skating competitions. By effectively assessing each action, our approach enhances the accuracy of TES predictions, thereby improving the precision and reliability of automated assessments in figure skating.
With Mamba’s superior ability to capture long-range dependencies \cite{yu2025mambaout}, our framework leverages Mamba's linear-time state space modeling to capture complete program contexts efficiently, which is ideal for handling long-time figure skating videos. 

The main contributions are summarized as follows:
\begin{itemize}
    \item We find that audio features are useless for TES evaluation. Thus, different from the existing figure skating score assessment methods, a two-stream network is presented for TES prediction with video contexts only and PCS prediction by video and audio fusion features. Both experiments and figure skating criteria prove that our framework is effective for figure skating score assessment.
    \item We propose a multi-scale Mamba pyramid network that accurately localizes, classifies, and scores action elements with the TES head at various temporal scales in figure skating. Compared to previous methods, our proposed method can predict the technical score for each action element in figure skating competitions.
    \item Our framework has the capability to capture long-range feature dependencies in long-time figure skating competitions by utilizing Mamba modules.  It achieves state-of-the-art performance on the FineFS benchmark and competitive results on two additional datasets without extra training, proving its effectiveness in robustness and transferability.
\end{itemize}

\section{Related Works}

\quad \textbf{Action Quality Assessment (AQA)} quantitatively evaluates human performance quality in sports, medical procedures, and artistic performances \cite{xu2024procedure,li2024continual,chen2024gaia,xu2024fineparser,zhou2024magr,dadashzadeh2024pecop,zhou2024cofinal}. It aims to simulate expert judgment by analyzing videos and assigning scores, a task challenged by subtle execution differences and subjective evaluation metrics \cite{tangUncertaintyAwareScoreDistribution2020,yuGroupawareContrastiveRegression2021}. Applications include sports training \cite{bruce2021skeleton}, surgical skill assessment \cite{gaoAsymmetricModelingAction2020}, and artistic performance evaluation \cite{zeng2020hybrid,liuFigureSkatingJumping2023}.
Early approaches used overall video representations for score regression \cite{parmarActionQualityAssessment2019}, but struggled with fine-grained details. Recent methods model temporal and spatial dynamics through hybrid models combining motion and posture features \cite{zeng2020hybrid}, and graph-based approaches for modeling body joint relationships \cite{leiMultiskeletonStructuresGraph2023,pan2019action}.
Current research focuses on integrating domain-specific knowledge like scoring rubrics for improved interpretability \cite{majeedi2024rica,okamoto2024hierarchical,xuLikertScoringGrade2022}, alongside developing fine-grained datasets such as FineDiving and FineFS \cite{xu2022finediving,liuFigureSkatingJumping2023}.

In figure skating AQA, approaches like the Replay-Guided method \cite{liuFigureSkatingJumping2023} use a Triple-Stream Contrastive Transformer with a Temporal Concentration Module to analyze multiple view angles, ensuring consistent feature learning across angles and zoom levels. It focuses on key segments rather than evenly distributed moments for scoring. Another approach \cite{liSkeletonBasedAction2021} uses skeletal data to evaluate key movements through three steps: extracting skeleton data via pose estimation, using a spatial-temporal graph convolution network (ST-GCN) to model dynamic skeletal changes, and predicting scores with a three-layer fully connected regressor. The field advances by integrating deep learning with domain-specific insights, enhancing both automated and expert assessments.
\begin{figure*}[t]
    \centering
    \includegraphics[width=\textwidth]{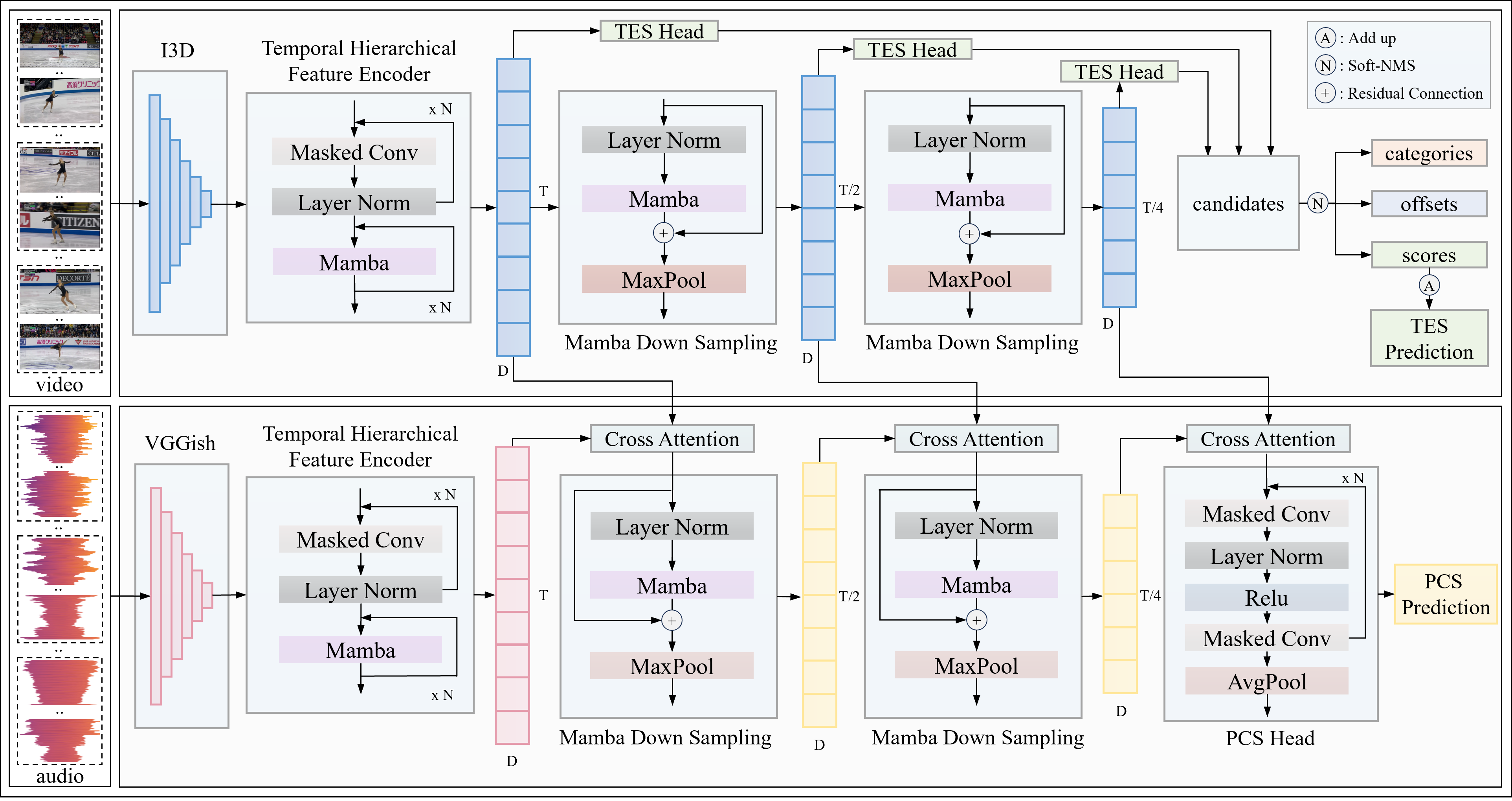} 
    \caption{Overview of our method. The proposed architecture contains visual context and visual-audio context streams. Temporal Hierarchical Feature Encoder (THFE) is used to capture hierarchical temporal features, followed by MDS (Mamba Down Sampling) modules to generate multi-scale Mamba pyramidal features. The visual context stream generates temporal offset, categories, and scores for each technical action element by TES Heads. The visual-audio context stream employs a parallel structure with similar components, interacting with the visual context stream via Cross Attention Fusion and finally regressing the PCS through the PCS Head.
    }
    \label{fig:overview}
\end{figure*}

\textbf{Temporal Action Localization (TAL)} plays an important role in Action Quality Assessment. AQA algorithms depend on TAL to identify and segment action instances before assessing their quality, particularly in sports and physical activities \cite{zhangLearningMotionRepresentation2020,heilbronActivityNetLargeScaleVideo2015,zhaoTemporalActionDetection2020,chaoRethinkingFasterRCNN2018}. In figure skating, diving and gymnastics, TAL locates specific movements, which are further evaluated for their execution quality.
LUSD-Net \cite{jiLocalizationassistedUncertaintyScore2023} employs an uncertainty score disentanglement approach to separate PCS-oriented and TES-oriented representations from skating sequences. For long-term sequences, they adopt a temporal interaction encoder to build contextual relationships between these different representations. It incorporates weakly-supervised temporal subaction localization to identify technical elements within extended performances. The approach demonstrates strong performance in scoring tasks, showcasing its potential for real-world sports analytics applications.
Leveraging TAL's capability to process untrimmed videos that encompass a range of action durations, AQA methods are enabled to align with human-defined evaluation criteria, thereby facilitating more precise and detailed assessments.

\textbf{Multi-modal learning} improves the accuracy and robustness of action evaluation in AQA. Traditional AQA systems primarily rely on visual information, such as RGB frames and optical flow, to assess the quality of human actions. However, incorporating additional modalities, such as audio or sensor data, can provide complementary cues that enhance the assessment process \cite{zengMultimodalActionQuality2024}. In figure skating and rhythmic gymnastics, background music and rhythmic patterns are essential for performance evaluation. Therefore, audio data can significantly enhance the accuracy of AQA systems.
For instance, Skating-Mixer \cite{xiaSkatingMixerLongTermSport2023} enhances the MLP framework with multi-modal inputs, effectively learning context representations via a memory recurrent unit (MRU), tailored for figure skating, focusing on movement and audio-visual coordination. Progressive Adaptive Multi-modal Fusion Network \cite{zengMultimodalActionQuality2024} processes RGB, optical flow, and audio through modality-specific and mixed-modality branches. It includes a Modality-specific Feature Decoder for selective information transfer, an Adaptive Fusion Module for tailored fusion policies, and a Cross-modal Feature Decoder for cross-modal feature integration. Semantics-Guided Network (SGN) \cite{duLearningSemanticsGuidedRepresentations2023} uses attention to facilitate adaptive knowledge transfer from semantics to visuals, employing learnable atomic queries in the student branch to mimic the teacher branch's semantic-aware distributions, aligning features across domains with three auxiliary losses. This multi-modal learning synergy is crucial for advancing AQA research, leading to more sophisticated action evaluation frameworks.

\section{Methodology}
The proposed architecture is illustrated in Figure \ref{fig:overview}. Our network integrates a visual context stream for TES prediction and a visual-audio context stream for PCS prediction.
The Temporal Hierarchical Feature Encoder (THFE) is utilized to capture hierarchical temporal features. This is followed by Mamba Down Sampling (MDS) modules that produce multi-scale pyramidal features. The visual context stream generates temporal offsets, categories, and scores for each technical action element through TES Heads. The visual-audio contexts stream processes audio input using a similar architecture. The two streams are connected via multi-level cross attention fusion that enables information exchange without influencing the visual context stream. Furthermore, the PCS is obtained from the visual-audio context stream through a specialized PCS Head that utilizes the fused features.

\subsection{Preliminary}
\textbf{Problem Formulation}
The purpose of the Action Quality Assessment of figure skating is to generate numerical scores from raw competition videos. The quality of a figure skater's competition performance is determined by a complex set of criteria. Understanding such complexity rules is essential for accurately assessing the true skill and artistry of the performance. 
TES is added by the score of each action element in one player's competition where each action is judged by action category and the quality of the action. TES has no relevance to audio contexts while lots of previous methods try to predict TES by visual and audio fusion features.
PCS contains 5 criteria which are Skating Skills, Transitions, Performance, Composition and Interpretation of the Music. PCS focuses on overall performance and needs video and audio contexts to help with the assessment.
Therefore, given a figure skating competition video and its corresponding audio, we should locate, classify and assess each action segment $P_i = (s_i, e_i, c_i, r_i) $, where $s_i \in [1, T]$, $e_i \in [1, T]$ is the start and end time point of each segment, and $c_i \in \{1,..., C\}$ is the class label, $r_i \in [0, a_{max}]$ is the score of the action, and $a_{max}$ denotes the maximum score for an individual action. We also need to get the PCS $r_{pcs} \in [0, pcs_{max}]$ based on the video audio input. The $pcs_{max}$ represents the maximum possible score for the overall program as defined by the competition rules.

\textbf{Feature Extractor}
Pre-trained I3D and VGGish are employed to derive feature vectors from both the video and its corresponding audio. These vectors are mapped to a unified embedding space through the application of two linear layers, generating video and audio features as $F_v, F_a \in \mathbb{R}^{T \times C} $, where $C$ represents the dimensionality of the embedding space and $T$ is the temporal length of the feature.

\subsection{Temporal Hierarchical Feature Encoder}
The Temporal Hierarchical Feature Encoder (THFE) is a novel component that is designed to capture hierarchical temporal features from video and audio embeddings and keep the features' temporal resolution. It consists of two key sub-modules: the Temporal Embedding Module (TEM) and the Temporal Refinement Module (TRM). These modules work sequentially to extract and refine temporal features, enabling robust representation learning for downstream tasks.

\paragraph{\textbf{Temporal Embedding Module (TEM)}}
The Temporal Embedding Module transforms the input features into a high-dimensional temporal embedding. Given an input feature sequence $\mathbf{F} \in \mathbb{R}^{ C \times T}$ from $F_v, F_a$, where $C$ is the input feature dimension, $T$ is the sequence length. TEM applies a series of 1D convolution layers, followed by ReLU activation and layer normalization. The $i$-th convolution layer in the TEM is formulated as:
\begin{equation}
\begin{aligned}
    \mathbf{F}_{i} = \text{ReLU}(\text{LayerNorm}(\text{Conv1D}(\mathbf{F}_{i-1}))),
\end{aligned}
\end{equation}
the output of the TEM is a feature sequence $\mathbf{F}_{\text{emb}} \in \mathbb{R}^{D \times T}$, where $D$ is the embedding dimension.

\begin{figure}[t]
    \centering
    \includegraphics[width=\linewidth]{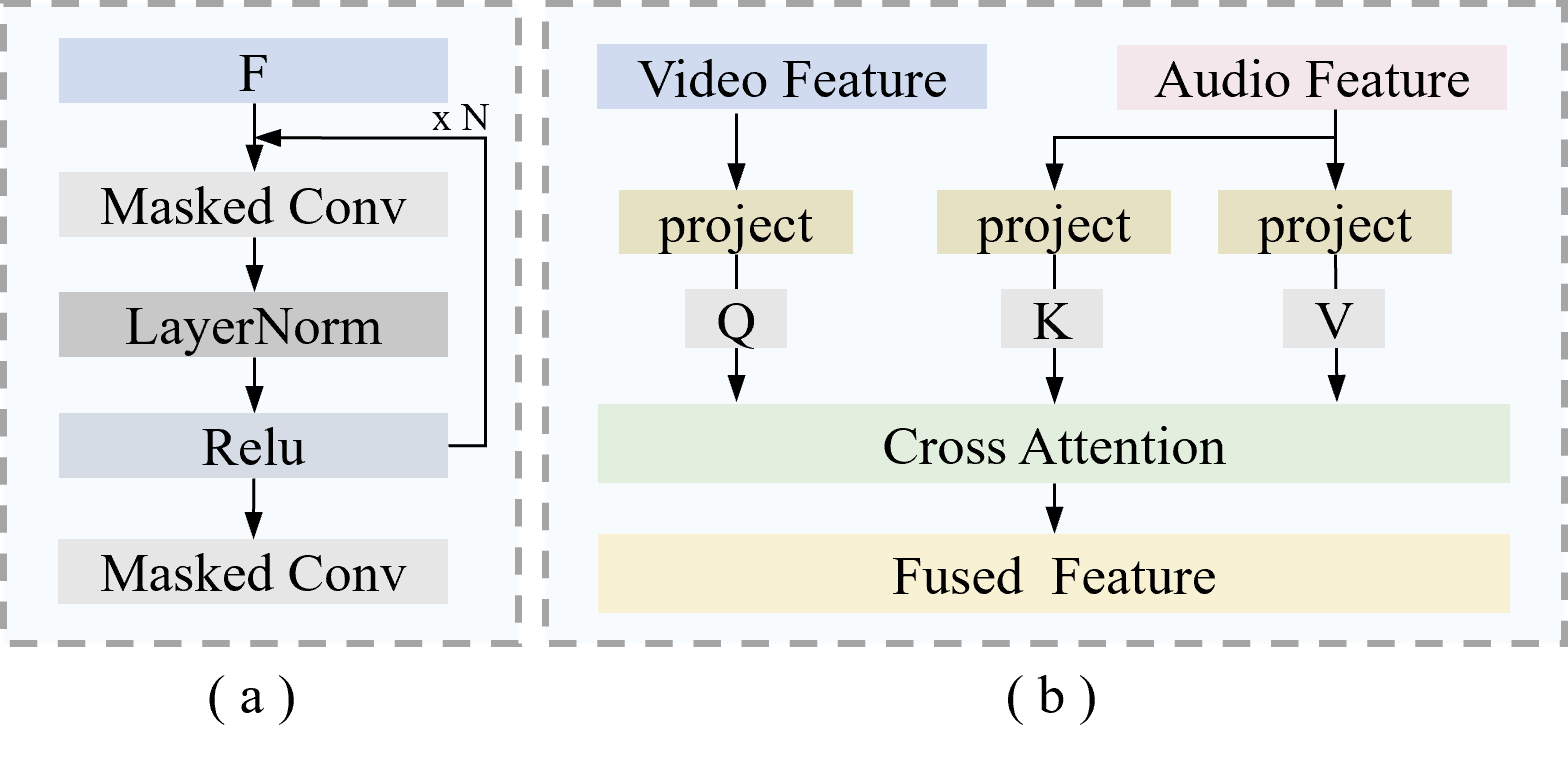} 
    \caption{(a) The TES head contains three similar structures with different parameters to classify, localize and score action elements at each time point. (b) The Cross Attention Module that fuses video and audio features.}
    \label{fig:mms}
\end{figure}

\paragraph{\textbf{Temporal Refinement Module (TRM)}}
The Temporal Refinement Module further encodes the temporal features using a series of Masked Mamba Blocks \cite{mamba2}. These blocks are designed to process the feature sequence while preserving the temporal structure through the masking operation. The TRM consists of $N$ such blocks. The output of the $j$-th Masked Mamba Block is computed as:
\begin{equation}
\begin{aligned}
    \mathbf{F}_{j} = \text{MaskMambaBlock}(\mathbf{F}_{j-1}, \mathbf{M}),
\end{aligned}
\end{equation}
where $\mathbf{M} \in \{0, 1\}^{T}$ is the binary mask indicating valid positions in the sequence. The output of the TRM is a refined feature sequence $\mathbf{F}_{\text{r}} \in \mathbb{R}^{D \times T}$, which retains the original temporal resolution.

\subsection{Multi-scale Mamba Pyramid}
The Multi-scale Mamba Pyramid (MMP) is a novel architecture designed to capture multi-scale temporal features by integrating Mamba Down Sampling (MDS) blocks. Each MDS block combines temporal feature extraction, residual connections and max pooling into a single unified operation, enabling hierarchical feature extraction at various temporal resolutions. This design is crucial for accurately assessing actions in figure skating since it captures both fine-grained details and high-level contextual information.

\paragraph{\textbf{Mamba Down Sampling (MDS) Block}}
The core component of the MMP is the Mamba Down sampling (MDS) block, which processes input features while preserving temporal structure through masking. Given the refined feature sequence $\mathbf{F_{r}} \in \mathbb{R}^{D \times T}$, the MDS block applies the following unified operation:
\begin{equation}
\begin{aligned}
    \mathbf{F}_{\text{m}} = \text{MaxPool}\left(\text{Mamba}\left(\text{LN}(\mathbf{F_{r}})\right) + \text{DropPath}(\mathbf{F_{r}}\right)),
\end{aligned}
\end{equation}
where $\text{LN}$ is the Layer Normalization operation, $\text{Mamba}$ is the Mamba-based temporal feature extraction module \cite{mamba2}, $\text{MaxPool}$ is the max-pooling operation with a configurable kernel size, stride, and padding, $\text{DropPath}$ is the stochastic depth regularization technique for residual connections.

The output feature sequence $\mathbf{F}_{\text{m}}$ is down-sampled and the corresponding mask $\mathbf{M}$ is implicitly handled by the Mamba module, ensuring that only valid positions are considered during pooling. 

The MMP stacks multiple MDS blocks with increasing down sampling strides, generating a pyramidal representation of features at different temporal resolutions. This hierarchical structure allows the model to capture both fine-grained details and high-level contextual information, which is essential for accurate action quality assessment. The outputs of the MMP are multi-scale feature sequences $\{\mathbf{F}^1, \mathbf{F}^2, \dots, \mathbf{F}^L\}$ and their corresponding masks $\{\mathbf{M}^1, \mathbf{M}^2, \dots, \mathbf{M}^L\}$, where $L$ is the layer number of the pyramid.

\paragraph{\textbf{Multi-level Cross Attention Fusion (MCAF)}}
As is shown in the Figure \ref{fig:mms} (b). The MCAF integrates multi-scale video feature sequences $\{\mathbf{F}^1_v, \mathbf{F}^2_v, \dots, \mathbf{F}^L_v\}$ and audio feature sequences $\{\mathbf{F}^1_a, \mathbf{F}^2_a, \dots, \mathbf{F}^L_a\}$ from the visual context and audio context streams, respectively. This module enables information flow from the visual context stream to the visual-audio context stream, ensuring that audio features enrich the visual representations without disrupting the scoring of technical elements. 

The fusion process operates hierarchically across multiple scales. At each level $l \in \{1, 2, \dots, L\}$, the video feature sequence $\mathbf{F}^l_v$ is used as the query \{$\mathbf{Q}^l$\} while the audio feature sequence $\mathbf{F}^l_a$ is used to generate the key \{$\mathbf{K}^l$\} and value \{$\mathbf{V}^l$\} projections. These projections are computed by using depthwise convolution layers followed by layer normalization: $\mathbf{Q}^l = \text{Conv1D}(\mathbf{F}^l_v)$, $\mathbf{K}^l = \text{Conv1D}(\mathbf{F}^l_a)$, and $\mathbf{V}^l = \text{Conv1D}(\mathbf{F}^l_a)$. The module then computes cross-attention between the video and audio features at each scale. The attention mechanism is formulated as:
\begin{equation}
\begin{aligned}
    \text{Attention}(\mathbf{Q}^l, \mathbf{K}^l, \mathbf{V}^l) = \text{Softmax}\left(\frac{\mathbf{Q}^l (\mathbf{K}^l)^T}{\sqrt{d_k}}\right) \mathbf{V}^l,
\end{aligned}
\end{equation}
where $d_k$ is the dimension of the key vectors. The attention scores are masked to ensure that only valid temporal positions are attended to. The output of the cross-attention mechanism is combined with the original video features through a residual connection, followed by a linear projection: $\mathbf{F}^l_{\text{fused}} = \text{Linear}(\text{Attention}(\mathbf{Q}^l, \mathbf{K}^l, \mathbf{V}^l)) + \mathbf{F}^l_v$. The fused features $\{\mathbf{F}^1_{\text{fused}}, \mathbf{F}^2_{\text{fused}}, \dots, \mathbf{F}^L_{\text{fused}}\}$ are then utilize to regress the PCS, ensuring enriched representations for accurate action quality assessment.

\subsection{Score Regression}
The score regression consists of TES Heads and a PCS Head. These heads produce the result for the figure skating TES and PCS assessment.

\textit{\textbf{TES Head}} is a unified module that predicts action categories, temporal offsets, and action scores for technical element assessment at each time point. As is shown in Figure \ref{fig:mms} (a), it consists of a series of 1D convolution layers followed by ReLU activation and layer normalization. The architecture is shared across the three tasks, with separate final layers for each output. Formally, for each feature level $l \in \{1, 2, \dots, L\}$, the intermediate features $\mathbf{H}^l$ are computed as:
\begin{equation}
\begin{aligned}
    \mathbf{H}^l = \text{ReLU}(\text{LayerNorm}(\text{Conv1D}(\mathbf{F}^l_v))),
\end{aligned}
\end{equation}
where $\mathbf{F}^l_v$ is the feature map at level $l$ of video pathway. The final outputs are computed using dedicated 1D convolution layers, where 
Action Categories: $\mathbf{C}^l = \text{Conv1D}_{\text{cls}}(\mathbf{H}^l)$, 
Temporal Offsets: $\mathbf{O}^l = \text{Conv1D}_{\text{reg}}(\mathbf{H}^l)$, 
Action Scores: $\mathbf{S}^l = \text{Conv1D}_{\text{score}}(\mathbf{H}^l)$,
where $\text{Conv1D}_{\text{cls}}$, $\text{Conv1D}_{\text{reg}}$, and $\text{Conv1D}_{\text{score}}$ are 1D convolution layers with output dimensions equal to the number of action categories, 2 (start and end offsets), and 1 (score per time point), respectively.

\textit{\textbf{PCS Head}}
predicts the overall PCS based on the fused features from the last level of the audio feature pyramid. As is shown in the Figure \ref{fig:overview}, it uses a series of 1D convolution layers followed by ReLU activation and layer normalization and finally applies an adaptive average pooling layer to produce a single score. Formally, the PCS score $\mathbf{r_{pcs}}$ is computed as: 
\begin{equation}
\begin{aligned}
    \mathbf{r_{pcs}} = \text{AvgPool}(\text{Conv1D}(\text{ReLU}(\text{LN}(\text{Conv1D}(\mathbf{F}^L_{fused}))))), 
\end{aligned}
\end{equation}
where $\mathbf{F}^L$ is the fused feature map at the last level $L$ in the visual-audio context stream. 

The final outputs of the score regression include TES outputs and PCS outputs. TES Head predicts action categories, temporal action offsets and action scores. PCS Head directly predicts PCS. These outputs collectively provide a comprehensive assessment of figure skating performance.

\subsection{Optimization}
\begin{figure}[t]
    \centering
    \includegraphics[width=\linewidth]{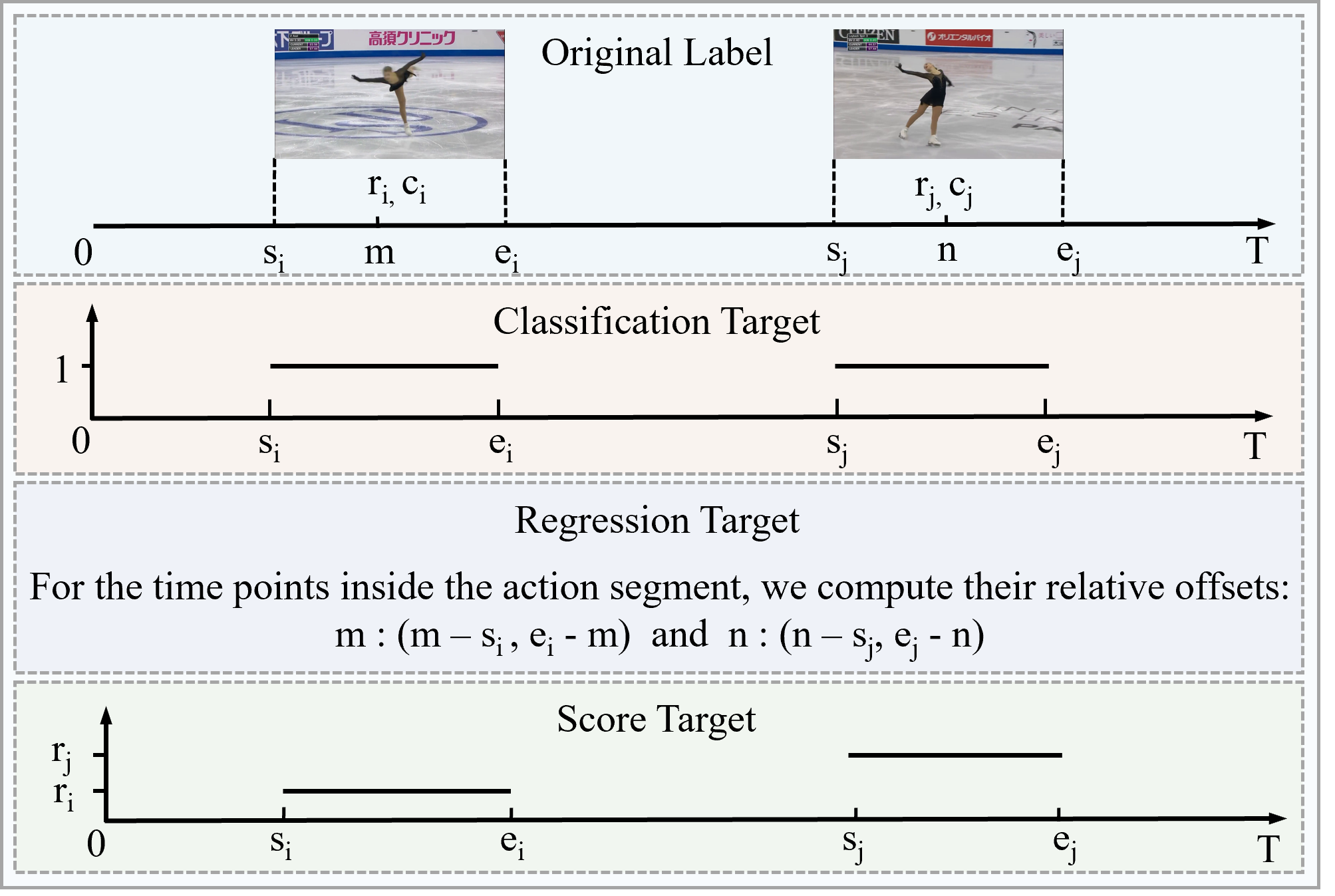}
    \caption{Point label generation for training.} 
    \label{fig:lg}
\end{figure}

We pad the temporal length of input features to a fixed length $T_{max}$. This ensures that the temporal dimension of each feature pyramid level is uniform for all samples. So, we can get a fixed number of time points at each level $\mathbf{P} = \{P_l\}_{l=1}^L$. In addition, we pre-define the regression range for the temporal offsets at each pyramid level, which allows us to normalize the regression targets across different levels.
As shown in Figure \ref{fig:lg}, the offset regression targets, along with action classification and score targets, are generated.


For action classification, one-hot encoding is applied to the time points within an action element, indicating the probability of a specific action at each moment.
For offset regression, we calculate the offset of the time point in the action element relative to the start and end of the action as the regression target.
The target setting processes of classification and offset follow \cite{zhang2022actionformer}.
For action element score regression, the value of the time point in the action element is assigned as the score for that action to obtain the score target.

\paragraph{\textbf{Loss Function}}
The optimization process involves minimizing a multi-task loss function that combines losses for action classification, temporal offset regression, element score prediction and PCS prediction. Specifically, we use the sigmoid focal loss $\mathcal{L}_{\text{focal}}$ \cite{lin2017focal} for action classification to handle class imbalance and improve the detection of rare actions. For temporal offset regression, we employ the differentiable IoU loss $\mathcal{L}_{\text{diou}}$ \cite{zheng2020distance} to directly optimize the overlap between predicted and ground truth segments. The mean squared error (MSE) loss $\mathcal{L}_{\text{element}}^{mse}$ is used to regress the quality scores for individual actions while the MSE loss $\mathcal{L}_{\text{pcs}}^{mse}$ is applied to predict the overall PCS.

The overall loss function is defined as:
\begin{equation}
\begin{aligned}
    \boldsymbol{\mathcal{L}} = \frac{\alpha \mathcal{L}_{\text{focal}} + \mathbb{I}_{positive} (\beta \mathcal{L}_{\text{diou}} + \mathcal{L}_{\text{element}}^{mse})}{N^{+}}  + \mathcal{L}_{\text{pcs}}^{mse},
\end{aligned}
\end{equation}
where $N^{+}$ is the total number of positive samples (time points within action segments and ground true offset inside pre-defined regress range), $\alpha$, $\beta$ are weighting coefficients that balance the contributions of the classification and offset regression, respectively. To enhance scoring performance, we don't implement weight balance on the element and PCS loss. The losses $\mathcal{L}_{\text{diou}}$ and $\mathcal{L}_{\text{element}}^{mse}$ are only applied to positive samples, as indicated by the indicator function $\mathbb{I}_{\text{positive}}$. The loss $\boldsymbol{\mathcal{L}}$ is computed for all levels of the feature pyramid in the mini-batch. This multi-task optimization ensures that the model learns to classify actions accurately, localize temporal boundaries, and predict quality scores simultaneously.

\section{Experiments and Results}
\subsection{Dataset and Evaluation Metrics}
\paragraph{\textbf{Dataset}}
To validate the effectiveness of our figure skating quality assessment model, we conduct comprehensive evaluations on three established benchmarks: Fis-V \cite{xuLearningScoreFigure2020}, FS1000 \cite{xiaSkatingMixerLongTermSport2023} and FineFS \cite{jiLocalizationassistedUncertaintyScore2023}. We only used the FineFS dataset for training since it has start time, end time and score of each technical action element.

The Fis-V dataset includes 500 videos of ladies' singles short programs, featuring 149 athletes from over 20 countries. In comparison, the FS1000 dataset is larger, with 1604 videos from elite competitions across all figure skating disciplines (singles, pairs, ice dance) and program types (short/free). The short program consists of 7 action elements, while the free skating includes 12 action elements. Each clip is trimmed to the performance segment ($\sim$3.5 min avg.) and split into 1247 training/validation and 357 test samples. The FineFS dataset provides a more detailed collection, with 1604 high-resolution videos ($\sim$5000 frames each at 25 FPS), annotated with precise action categories, segments and scores. It includes 1167 performances (729 short programs, 438 free skates) and maintains a balanced gender distribution (570 male, 597 female athletes). For consistency, all datasets retain their original train-test splits.

\paragraph{\textbf{Evaluation Metrics}} To maintain alignment with standard practices in the field, we adopt Spearman's rank correlation coefficient (abbreviated as SP $\rho$) as our key evaluation metric. This measure, which varies between -1 and 1, assesses the strength of the relationship between predicted scores and ground-truth annotations. A higher coefficient signifies a greater correspondence between the two sets of rankings. The mathematical expression for this metric is given as follows:
\begin{equation}
\begin{aligned}
    \rho=\frac{\sum_i(a_i^r-\bar{b}^r)(b_i^r-\bar{b}^r)}{\sqrt{\sum_i(a_i^r-\bar{b}^r)^2\sum_i(b_i^r-\bar{b}^r)^2}},
\end{aligned}
\end{equation}
where $a^r$ and $b^r$ represent the rank positions of two distinct series, while $\bar{a}^r$, $\bar{b}^r$ denote the mean rank values across all samples in series a and b, respectively. The Spearman correlation coefficient $\rho$ spans from -1 to 1, quantifying the degree of alignment in ranking order between the predicted outcomes and the ground truth. A higher value of $\rho$ signifies a stronger correlation and better performance.

Furthermore, to assess the performance of action localization within our study, we employ the mean average precision (mAP) with different tIOU thresholds as the metric of evaluation. Specifically, we apply the mAP@[0.5 : 0.05 : 0.95] across all datasets and present the average mAP as our results.

\begin{table}[t]
\centering
\caption{Comparison with state-of-the-art approaches on FineFS dataset.}
\begin{tabular}{c|cc|cc}
\toprule 
\multirow{2}{*}{Method} & \multicolumn{2}{c|}{Free Skating} & \multicolumn{2}{c}{Short Program} \\
\cmidrule{2-5}
   & $\rho_{TES}$ & $\rho_{PCS}$ & $\rho_{TES}$ & $\rho_{PCS}$ \\
\midrule 
GDLT\cite{xuLikertScoringGrade2022} & 0.63 & 0.80 & 0.64 & 0.78\\
MS-LSTM\cite{xuLearningScoreFigure2020} & 0.55 & 0.62 & 0.55 & 0.60\\
TSA\cite{xu2022finediving} & 0.68 & 0.78 & 0.53 & 0.78 \\
LUSD-Net \cite{jiLocalizationassistedUncertaintyScore2023} & 0.78 & 0.86 & 0.69 & 0.81 \\
\textbf{Ours} & \textbf{0.80} & \textbf{0.96} & \textbf{0.75} & \textbf{0.94} \\
\bottomrule 
\end{tabular}
\label{tab:finefs}
\end{table}

\subsection{Implementation Details}
Initially, we input 24 consecutive frames at a frequency of 24 frames per second into an I3D model \cite{carreira2017quo}, pre-trained on the Kinetics-400 dataset \cite{kay2017kinetics}, for the extraction of RGB features. Concurrently, audio features are extracted utilizing the VGGish model \cite{hershey2017cnn}, which is pre-trained on the AudioSet dataset \cite{gemmeke2017audio}, with feature extraction occurring every second to achieve temporal alignment with the RGB features. In addition, parameter $T_{max}$ is configured to 288 in accordance with the competition show's duration, and a 22-category classification strategy is adopted to optimize the technical element scoring based on \cite{liuTemporalSegmentationFinegained2021}. A six-level feature pyramid is employed with video and audio features fused at each level. 
The regress range for each pyramid level is: [0,4], [4,8], [8,16], [16,32], [32,64], [64,$+\infty$], which follows \cite{zhang2022actionformer}. These ranges are used to define positive samples. 
The model is trained over 50 epochs with a batch size of 8 on an NVIDIA A800 GPU, using an initial learning rate of $1e-3$ and an Adam optimizer with a weight decay of $5e-2$. A learning rate scheduler is utilized, diminishing the rate by 0.1 every 5 epochs.
Furthermore, we employ linear layers to project video and audio features to a uniform dimension of 1024.
During inference, our model outputs $(s_i, e_i, c_i, r_i) $ for each time point $i$ and $r_{pcs}$ for the whole sample; the action segments undergo further processing using soft-NMS \cite{neubeck2006efficient} to remove segments that overlap significantly, and the iou threshold is 0.1.
This process yields the action segments along with their associated categories and scores. 
For the short program samples, we select the top seven segments based on category probabilities, while for the free skating samples, we choose the top twelve segments using the same criterion. 
These selected segments constitute the final output. Lastly, we sum the element scores of each action segment to compute the TES for the sample.

\begin{table}[t]
\centering
\caption{Comparison with state-of-the-art approaches on the other datasets (without training).}
\begin{tabular}{c|c|cc}
\midrule
Dataset & Method  &  $\rho_{TES}$ & $\rho_{PCS}$ \\
\midrule
\multirow{8}{*}{Fis-V}  & MS-LSTM \cite{xuLearningScoreFigure2020} & 0.65 & 0.78   \\
                      & M-BERT \cite{lee2021parameter} & 0.68 & 0.82   \\
                      & Action-Net \cite{zeng2020hybrid} & \textbf{0.81} & 0.70  \\                      
                      & Skating-Mixer \cite{xiaSkatingMixerLongTermSport2023} & 0.68 & 0.82   \\
                      & Semantic-Guide \cite{duLearningSemanticsGuidedRepresentations2023} & 0.70 & 0.83  \\
                      & PAMFN \cite{zengMultimodalActionQuality2024} & 0.79 & \textbf{0.89} \\
                      & GDLT \cite{xuLikertScoringGrade2022} & 0.69 & 0.82  \\
                      & LUSD-Net \cite{jiLocalizationassistedUncertaintyScore2023} & 0.68 & 0.82 \\
\cmidrule{2-4}
                      & Ours & 0.79 & \textbf{0.87}  \\
\midrule
\multirow{5}{*}{FS1000} & MS-LSTM \cite{xuLearningScoreFigure2020} & 0.86 & 0.80   \\
                      & M-BERT \cite{lee2021parameter} & 0.79 & 0.75   \\
                      & Skating-Mixer \cite{xiaSkatingMixerLongTermSport2023} & 0.88 & 0.82  \\
                      & Semantic-Guide \cite{duLearningSemanticsGuidedRepresentations2023} & \textbf{0.89} & 0.85 \\
\cmidrule{2-4}
                      & Ours & 0.85 & \textbf{0.91} \\
\midrule
\end{tabular}
\label{tab:fisvfs1000}
\end{table}

\subsection{Comparison with State-of-the-Art Methods}

\begin{table*}[t]
\centering
\caption{Evaluation on the number of action categories.}
\begin{tabular}{c|cccccccccc|c|cc}
\toprule
\multirow{2}{*}{Number of Categories} & \multicolumn{11}{c|}{Temporal Intersection Over Union (tIOU)} & \multicolumn{2}{c}{SP} \\
\cmidrule{2-14}
  & 0.50 & 0.55 & 0.60 & 0.65 & 0.70 & 0.75 & 0.80 & 0.85 & 0.90 & 0.95 & Avg & TES & PCS \\
\midrule
4 categories & 92.51 & 90.99 & 89.23 & 86.74 & 81.64 & 75.53 & 70.00 & 63.19 & 53.40 & 33.04 & 73.63 & 0.73 & 0.93 \\
8 categories & 94.86 & 92.66 & 91.34 & 88.50 & 84.29 & 79.30 & 71.41 & 64.27 & 54.77 & 33.82 & 75.52 & 0.67 & 0.93\\
\textbf{22 categories} & 81.01 & 79.49 & 78.81 & 77.68 & 74.06 & 70.73 & 65.35 & 60.60 & 53.21 & 34.41 & 67.53  & \textbf{0.77} & \textbf{0.95}\\
242 categories & 46.55 & 45.96 & 45.37 & 44.59 & 43.12 & 41.31 & 38.96 & 36.61 & 31.28 & 17.91 & 39.17  & 0.72 & 0.93\\

\bottomrule
\end{tabular}
\label{tab:category}
\end{table*}

We evaluate our proposed model on the FineFS dataset and compare its performance against state-of-the-art methods. As is shown in Table \ref{tab:finefs}, our model achieves superior performance across all metrics, with $\rho_{TES}$ and $\rho_{PCS}$ scores of 0.80 and 0.96 for Free Skating, and 0.75 and 0.94 for Short Program, respectively. These results surpass the previous best results, demonstrating the effectiveness of our approach.

A key advantage of our method is the incorporation of audio modality, which is not used by any of the compared approaches. By integrating audio features extracted using the VGGish model \cite{hershey2017cnn}, our model captures critical information related to music interpretation and synchronization, enhancing the assessment of program components. This multi-modal fusion strategy, combined with hierarchical feature extraction and score regression, enables our model to achieve state-of-the-art performance, particularly in PCS evaluation. The improvements highlight the importance of leveraging both visual and audio modalities for comprehensive figure skating quality assessment.

Moreover, the fine-grained action assessment leads to better TES prediction, as our multi-scale motion representation strategy enables precise localization and evaluation of individual skating elements, capturing subtle technical details that are critical for accurate TES assessment.

We evaluate our model's generalization capability on the Fis-V and FS1000 datasets without additional training. As is shown in Table \ref{tab:fisvfs1000}, our model achieves competitive performance on both datasets, demonstrating strong transferability of the proposed method. On FS1000, our method excels other methods in PCS evaluation, indicating the benefits of incorporating audio modality. 

\begin{table}[t]
\centering
\caption{Comparison on temporal encoder structure through the replacement of the Mamba block in our method.}
\begin{tabular}{c|c|ccc|cc}
\toprule
\multirow{2}{*}{Structure} & \multirow{2}{*}{Params} & \multicolumn{3}{c|}{tIOU} & \multicolumn{2}{c}{SP} \\
\cmidrule{3-7}
  & & 0.50 & 0.75 & 0.95 &  TES & PCS \\
\midrule
Conv(400) & 40.23M & 63.88 & 49.63 & 15.08 &  0.71 & 0.80\\
Conv(512) & 64.24M & 64.27 & 50.12 & 14.68 &  0.68 & 0.76\\
\textbf{Mamba(512)} & \textbf{40.45M} &\textbf{81.01} & \textbf{70.73} & \textbf{34.41} & \textbf{0.77} & \textbf{0.95}\\
\bottomrule
\end{tabular}
\label{tab:com}
\end{table}

\subsection{Ablation Study}
We conduct ablation studies on the FineFS dataset 
which provides comprehensive temporal annotations, as well as detailed categorizations and scores for individual movements.

\begin{table}[t]
\centering
\caption{Evaluation on the number of layers in the Mamba pyramid and the range for initialization regression.}
\begin{tabular}{c|c|ccc|c|cc}
\toprule
\multirow{2}{*}{Levels} & \multirow{2}{*}{Init} & \multicolumn{4}{c|}{tIOU} & \multicolumn{2}{c}{SP} \\
\cmidrule{3-8}
 &  & 0.50 & 0.75 & 0.95 & Avg & TES & PCS \\
\midrule
1 & [0,$+\infty$) & 78.07 & 67.21 & 25.55 & 63.42  & 0.72 & 0.84\\
4 & [0,4) & 79.04 & 68.32 & 26.42 & 62.24  & 0.73 & 0.89\\
7 & [0,4) & 79.41 & 69.54 & 28.64 & 64.82  & 0.74 & 0.93\\
\midrule
6 & [0,2) & 80.64 & 69.62 & \textbf{35.24} & 65.42  & 0.76 & 0.94\\
\textbf{6} & \textbf{[0,4)} & \textbf{81.01} & \textbf{70.73} & 34.41 & \textbf{67.53} & \textbf{0.77} & \textbf{0.95}\\
6 & [0,8) & 76.99 & 68.43 & 26.43 & 64.19  & 0.76 & 0.93\\
\bottomrule
\end{tabular}
\label{tab:level}
\end{table}

\textbf{Evaluation on the number of action categories.}
We conduct an ablation study to investigate the impact of category numbers on temporal action localization and score prediction. Specifically, we evaluate four settings with varying numbers of action categories: 4, 8, 22, and 242. The results, presented in Table \ref{tab:category}, demonstrate that finer-grained category definitions (e.g., 22 categories) achieve the best balance between temporal localization accuracy and score prediction performance. While coarser granularity (e.g., 4 or 8 categories) yields higher mAP for temporal localization, it sacrifices precision in TES prediction. Conversely, overly fine-grained granularity (e.g., 242 categories) significantly degrades localization performance due to increased complexity. The 22-category setting achieves the highest Spearman correlation for both TES and PCS, highlighting its effectiveness in capturing the nuances of figure skating actions while maintaining robust localization capabilities. This study underscores the importance of selecting an appropriate number of action categories for optimal performance in action quality assessment.

\textbf{Comparison on temporal encoder structures.}
To demonstrate the performance of the Mamba module, we compared it with another temporal encoder approach by replacing the Mamba blocks with CNN. For a fair comparison, we reduced the dimensions of all intermediate features in the CNN architecture from 512 to 400, keeping the total parameter count close to that of the Mamba architecture. As is shown in Table~\ref{tab:com}, Mamba outperforms CNN on all metrics, regardless of the number of parameters used.
This demonstrates Mamba's exceptional suitability for evaluating figure skating performances.

\textbf{Evaluation on the number of layers in the Mamba pyramid and the range for initialization regression.}
We conduct an ablation study to evaluate the impact of different feature pyramid layers and initialization ranges on temporal localization and score prediction. According to our analysis, the temporal duration of one action ranges from 1 to 77 seconds. The results, presented in Table \ref{tab:level}, demonstrate that using 6 pyramid levels with an initialization range of [0,4) achieves the best performance.
This configuration strikes an optimal balance between capturing fine-grained details and high-level contextual information, enabling precise temporal localization and accurate score prediction. The study highlights the importance of carefully selecting the number of pyramid levels and initialization ranges to achieve robust performance in figure skating quality assessment.

\begin{table}[t]
\centering
\caption{Comparison on audio integration strategies.}
\begin{tabular}{c|ccc|cc}
\toprule
\multirow{2}{*}{Integration Method} & \multicolumn{3}{c|}{tIOU} & \multicolumn{2}{c}{SP} \\
\cmidrule{2-6}
 & 0.50 & 0.75 & 0.95 & TES & PCS \\
\midrule
w/o Audio & 70.53 & 63.34  & 28.63 & 0.75 & 0.89\\
Symmetrical Fusion & 75.56 & 65.84 & 26.52 & 0.73 & 0.94 \\
One Stream Fusion & 56.29 & 49.63 & 15.37  & 0.70 & 0.75 \\
\textbf{Two Stream Fusion} & \textbf{81.01} & \textbf{70.73} & \textbf{34.41} & \textbf{0.77} & \textbf{0.95}\\
\bottomrule
\end{tabular}
\label{tab:audio}
\end{table}

\textbf{Comparison on audio integration strategies.}
We conduct a systematic evaluation of different audio integration strategies. We compare the performance between using audio and visual features and only using visual features for the tasks. 
One stream fusion uses the fused audio and video features to obtain TES and PCS at the same time. 
We also add a symmetrical two-stream fusion baseline, where audio features are fused into the TES branch via cross-attention for fairer comparison.
Our proposed two-stream method selectively incorporates audio cues only for PCS prediction. As is shown in Table \ref{tab:audio}, our approach achieves superior performance across all metrics, demonstrating three key conclusions:
Firstly, the video-only and symmetrical two-stream fusion approach shows that audio features are non-essential for TES prediction, but beneficial for PCS; secondly, one stream fusion severely degrades both TES and localization performance, confirming that using the fused feature to obtain both TES and PCS will harm the performance; Thirdly, our two-stream fusion method preserves TES accuracy while significantly boosting PCS prediction, validating our design of modality-specific feature processing.

\begin{table}[t]
\centering
\caption{Comparison on MCAF fusion levels.}
\begin{tabular}{c|c|c}
\toprule
\multirow{2}{*}{Fusion Level} & \multirow{2}{*}{Fusion Times} & SP \\
\cmidrule{3-3}
 & & PCS \\
\midrule
At Level 1 & 1 &  0.9183\\
At Level 3 & 1 & 0.9252\\
At Level 5 & 1 & 0.9435\\
\midrule
At Levels 1, 3, 5 & 3 & 0.9283\\
\textbf{At Levels 1 $\sim$ 6} & 6 & \textbf{0.9526}\\
\bottomrule
\end{tabular}
\label{tab:fpnfusion}
\end{table}

\textbf{Comparison on MCAF fusion levels.}
We conduct an ablation study to evaluate the impact of different MCAF fusion levels on PCS prediction. The results, presented in Table \ref{tab:fpnfusion}, demonstrate that fusing features across all levels (Levels 1 $\sim$ 6) achieves the highest Spearman correlation, outperforming fusion at individual levels or partial combinations. This indicates that leveraging multi-scale features from all MCAF levels provides a more comprehensive representation, capturing both fine-grained details and high-level contextual information essential for accurate PCS prediction. The study highlights the importance of multi-scale feature fusion in our framework and underscores its effectiveness in enhancing the model's ability to assess artistic and interpretative aspects of figure skating performances.

\textbf{Comparison on the Query Type in Cross-Attention.}
We conduct an ablation study to evaluate the impact of query type in cross-attention for PCS prediction. 
Specifically, two configurations are compared: one case is video features are projected as query features while audio features are projected as key and value features, and the converse scenario.
The results, presented in Table \ref{tab:query}, demonstrate that using video as query achieves the highest Spearman correlation, outperforming using audio as query.

Using video as query allows the model to leverage visual features as the primary source of information while incorporating audio features as supplementary context, leading to more accurate PCS predictions. This study underscores the effectiveness of our cross-attention mechanism and the critical role of multi-modal fusion in achieving state-of-the-art PCS prediction.

\textbf{Evaluation on loss weight configuration.}
We evaluate the impact of loss weights ($\alpha$ for $\mathcal{L}_{\text{focal}}$, $\beta$ for $\mathcal{L}_{\text{diou}}$) on multi-task learning. The best configuration ($\alpha = 0.7$, $\beta = 0.3$) achieves superior tIOU and Spearman scores, demonstrating the importance of balanced weighting for classification and localization, as it enhances the model's ability to distinguish between fine-grained action categories and assess their quality.

\begin{table}[t]
\centering
\caption{Comparison on the Query Type in Cross-Attention.}
\begin{tabular}{c|cc|c}
\toprule
\multirow{2}{*}{Query} & \multirow{2}{*}{Key} & \multirow{2}{*}{Value} & SP \\
\cmidrule{4-4}
& & & PCS \\
\midrule
Audio & \multicolumn{2}{c|}{Video}  & 0.9359\\
Video & \multicolumn{2}{c|}{Audio}  & 0.9526\\
\bottomrule
\end{tabular}
\label{tab:query}
\end{table}

\begin{table}[t]
\centering
\caption{Evaluation on loss weight configuration.}
\begin{tabular}{cc|ccc|c|ccc}
\toprule
\multicolumn{2}{c|}{Weight} & \multicolumn{4}{c|}{tIOU} & \multicolumn{2}{c}{SP} \\
\midrule
$\alpha$ & $\beta$ & 0.50 & 0.75 & 0.95 & Avg & TES & PCS \\
\midrule
0.5 & 0.5 & 75.18 & 66.42  & 31.82 & 63.55  & 0.76 & 0.92\\
0.3 & 0.7 & 78.41 & 68.53 & 32.79 & 65.73 & 0.75 & 0.93 \\
\textbf{0.7} & \textbf{0.3} & \textbf{81.01} & \textbf{70.73} & \textbf{34.41} & \textbf{67.53} & \textbf{0.77} & \textbf{0.95}\\
\bottomrule
\end{tabular}
\label{tab:weight}
\end{table}

\section{Conclusion}
This paper introduces a two-stream mamba pyramid network that accurately assesses the performance of figure skaters in competitions. We propose an approach that aligns with actual judging criteria by separating visual-feature based TES evaluation stream from audio-visual-feature based PCS evaluation stream. Our multi-scale Mamba pyramid and TES head effectively address the challenges of localizing and assessing action elements with various temporal scales in figure skating. The multi-level fusion mechanism ensures that video-based features are unaffected when assessing TES and enhances PCS estimation by integrating visual and auditory cues across each level of the pyramid. With Mamba’s superior ability to capture long-range dependencies and its linear computational complexity, our method is ideally suited for handling lengthy figure skating videos.
Experimental results confirm our framework's effectiveness, achieving state-of-the-art performance on the FineFS benchmark and demonstrating strong transferability across two additional datasets without extra training. 
\section{Acknowledgments}
This work is supported by the National Key Research and Development Program of China (Grant No. 2022YFC3600403) and Emerging Interdisciplinary Platform for Medicine and Engineering in Sports (EIPMES), Beijing, China.

\bibliographystyle{ACM-Reference-Format}
\balance
\bibliography{main}


\onecolumn
\appendix
\renewcommand\thetable{\Alph{section}\arabic{table}}
\section{The architecture Details}
\setcounter{table}{0}

As is shown in Figure \ref{fig:as}, we illustrate the layers within the visual context and visual-audio context streams, including the accompanying TES and PCS Head, explaining their inputs and outputs. In convolution layers, \( k \) denotes the kernel size for 1D convolutions, while \( s \) represents the stride. The variables \( c_i \) and \( c_o \) stand for the input and output feature channels, respectively. The term \( ds \) indicates the downsampling ratio. \( T \) is the temporal length of the input sequence, and \( D \) signifies the input feature dimension. For classification within the TES Head, the output dimension \( N \) is set to 22, the output dimension \( N \) equals 2 for offset regression and 1 for action element score regression.

\begin{table*}[!ht]

\centering
\caption{The detailed info about our architecture.}
\label{fig:as}
\begin{tabular}{c|c|c|c|c}
\toprule
 & Name & Input & Layers & Output Size \\

\midrule

\multirow{10}{*}{Mamba(V)} & video\_feature & video & I3D & T$\times$D \\
\cmidrule{2-5}
&\multirow{4}{*}{V\_THFE} & \multirow{4}{*}{video\_feature} & conv $k$=3, $s$=1 ($c_i$ = D, $c_o$ = 512) & T $\times$ 512 \\
& &  & conv $k$=3, $s$=1 ($c_i$ = 512, $c_o$ = 512) & T $\times$ 512 \\
& &  & Mamba & T $\times$ 512 \\
& &  & Mamba & T $\times$ 512 \\
\cmidrule{2-5}
& V\_MDS 1 & V\_THFE & Mamba Down Sample Unit, $ds$=2 & T/2 $\times$512 \\
& V\_MDS 2 & V\_MDS 1 & Mamba Down Sample Unit, $ds$=2 & T/4 $\times$512 \\
& V\_MDS 3 & V\_MDS 2 & Mamba Down Sample Unit, $ds$=2 & T/8 $\times$512 \\
& V\_MDS 4 & V\_MDS 3 & Mamba Down Sample Unit, $ds$=2 & T/16 $\times$512 \\
& V\_MDS 5 & V\_MDS 4 & Mamba Down Sample Unit, $ds$=2 & T/32 $\times$512 \\
\midrule

\multirow{3}{*}{TES Head} & \multirow{3}{*}{cls/reg/score} & \multirow{3}{*}{V\_THFE,V\_MDS 1,...,V\_MDS 5} & conv $k$=3, $s$=1 ($c_i$ = 512, $c_o$ = 512) & [T x 512,...T/32 x 512] \\
& &  & conv $k$=3, $s$=1 ($c_i$ = 512, $c_o$ = 512) & [T x 512,...T/32 x 512] \\
& &  & conv $k$=3, $s$=1 ($c_i$ = 512, $c_o$ = N) & [T x N,...T/32 x N] \\

\midrule
\multirow{10}{*}{Mamba(VA)} & audio\_feature & audio & Vggish & T$\times$D \\
\cmidrule{2-5}
&\multirow{4}{*}{A\_THFE} & \multirow{4}{*}{audio\_feature} & conv $k$=3, $s$=1 ($c_i$ = D, $c_o$ = 512) & T $\times$ 512 \\
& &  & conv $k$=3, $s$=1 ($c_i$ = 512, $c_o$ = 512) & T $\times$ 512 \\
& &  & Mamba & T $\times$ 512 \\
& &  & Mamba & T $\times$ 512 \\
\cmidrule{2-5}
& VA\_MDS 1 & cross\_attention(V\_THFE,A\_THFE) & Mamba Down Sample Unit, $ds$=2 & T/2 $\times$512 \\
& VA\_MDS 2 & cross\_attention(VA\_MDS 1,V\_MDS 1) & Mamba Down Sample Unit, $ds$=2 & T/4 $\times$512 \\
& VA\_MDS 3 & cross\_attention(VA\_MDS 2,V\_MDS 2) & Mamba Down Sample Unit, $ds$=2 & T/8 $\times$512 \\
& VA\_MDS 4 & cross\_attention(VA\_MDS 3,V\_MDS 3) & Mamba Down Sample Unit, $ds$=2 & T/16 $\times$512 \\
& VA\_MDS 5 & cross\_attention(VA\_MDS 4,V\_MDS 4) & Mamba Down Sample Unit, $ds$=2 & T/32 $\times$512 \\
\midrule

\multirow{4}{*}{PCS Head} & \multirow{4}{*}{PCS score} & \multirow{4}{*}{VA\_MDS 5} & conv $k$=3, $s$=1 ($c_i$ = 512, $c_o$ = 512) & [T/32 x 512] \\
& &  & conv $k$=3, $s$=1 ($c_i$ = 512, $c_o$ = 512) & [T/32 x 512] \\
& &  & conv $k$=3, $s$=1 ($c_i$ = 512, $c_o$ = 1) & [T/32 x 1] \\
& &  & Average Pooling & 1 \\
\bottomrule
\end{tabular}
\end{table*}

\newpage
As is shown in Figure \ref{fig:a}, we provide detailed information on the convolution architecture used in the temporal encoder structure ablation study. In summary, we replace the Mamba block in our architecture with convolution layers. The meanings of each symbol are consistent with those in Figure \ref{fig:as}. To ensure a fair comparison, the dimensions of all intermediate features were reduced from 512 to 400. This adjustment was made to keep the total number of model parameters close to that used with Mamba.

\begin{table*}[!ht]
\centering
\caption{The conv architecture used in the temporal encoder structure ablation study.}
\label{fig:a}
\begin{tabular}{c|c|c|c|c}
\toprule
 & Name & Input & Layers & Output Size \\
\midrule
\multirow{10}{*}{Conv(V)} & video\_feature & video & I3D & T$\times$D \\
\cmidrule{2-5}
&\multirow{4}{*}{V\_THFE} & \multirow{4}{*}{video\_feature} & conv $k$=3, $s$=1 ($c_i$ = D, $c_o$ = 512) & T $\times$ 512 \\
& &  & conv $k$=3, $s$=1 ($c_i$ = 512, $c_o$ = 512) & T $\times$ 512 \\
& &  & conv $k$=3, $s$=1 ($c_i$ = 512, $c_o$ = 512) & T $\times$ 512 \\
& &  & conv $k$=3, $s$=1 ($c_i$ = 512, $c_o$ = 512) & T $\times$ 512 \\
\cmidrule{2-5}
& V\_CDS 1 & V\_THFE & Conv Down Sample Unit, $k$=3, $s$=1, $ds$=2 & T/2 $\times$512 \\
& V\_CDS 2 & V\_CDS 1 & Conv Down Sample Unit, $k$=3, $s$=1, $ds$=2 & T/4 $\times$512 \\
& V\_CDS 3 & V\_CDS 2 & Conv Down Sample Unit, $k$=3, $s$=1, $ds$=2 & T/8 $\times$512 \\
& V\_CDS 4 & V\_CDS 3 & Conv Down Sample Unit, $k$=3, $s$=1, $ds$=2 & T/16 $\times$512 \\
& V\_CDS 5 & V\_CDS 4 & Conv Down Sample Unit, $k$=3, $s$=1, $ds$=2 & T/32 $\times$512 \\
\midrule
\multirow{3}{*}{TES Head} & \multirow{3}{*}{cls/reg/score} & \multirow{3}{*}{V\_THFE,V\_CDS 1,...,V\_CDS 5} & conv $k$=3, $s$=1 ($c_i$ = 512, $c_o$ = 512) & [T x 512,...T/32 x 512] \\
& &  & conv $k$=3, $s$=1 ($c_i$ = 512, $c_o$ = 512) & [T x 512,...T/32 x 512] \\
& &  & conv $k$=3, $s$=1 ($c_i$ = 512, $c_o$ = N) & [T x N,...T/32 x N] \\

\midrule
\multirow{10}{*}{Conv(VA)} & audio\_feature & audio & Vggish & T$\times$D \\
\cmidrule{2-5}
&\multirow{4}{*}{A\_THFE} & \multirow{4}{*}{audio\_feature} & conv $k$=3, $s$=1 ($c_i$ = D, $c_o$ = 512) & T $\times$ 512 \\
& &  & conv $k$=3, $s$=1 ($c_i$ = 512, $c_o$ = 512) & T $\times$ 512 \\
& &  & conv $k$=3, $s$=1 ($c_i$ = 512, $c_o$ = 512) & T $\times$ 512 \\
& &  & conv $k$=3, $s$=1 ($c_i$ = 512, $c_o$ = 512) & T $\times$ 512 \\
\cmidrule{2-5}
& VA\_CDS 1 & cross\_attention(V\_THFE,A\_THFE) & Conv Down Sample Unit, $k$=3, $s$=1, $ds$=2 & T/2 $\times$512 \\
& VA\_CDS 2 & cross\_attention(VA\_CDS 1,V\_CDS 1) & Conv Down Sample Unit, $k$=3, $s$=1, $ds$=2 & T/4 $\times$512 \\
& VA\_CDS 3 & cross\_attention(VA\_CDS 2,V\_CDS 2) & Conv Down Sample Unit, $k$=3, $s$=1, $ds$=2 & T/8 $\times$512 \\
& VA\_CDS 4 & cross\_attention(VA\_CDS 3,V\_CDS 3) & Conv Down Sample Unit, $k$=3, $s$=1, $ds$=2 & T/16 $\times$512 \\
& VA\_CDS 5 & cross\_attention(VA\_CDS 4,V\_CDS 4) & Conv Down Sample Unit, $k$=3, $s$=1, $ds$=2 & T/32 $\times$512 \\
\midrule
\multirow{4}{*}{PCS Head} & \multirow{4}{*}{PCS score} & \multirow{4}{*}{VA\_CDS 5} & conv $k$=3, $s$=1 ($c_i$ = 512, $c_o$ = 512) & [T/32 x 512] \\
& &  & conv $k$=3, $s$=1 ($c_i$ = 512, $c_o$ = 512) & [T/32 x 512] \\
& &  & conv $k$=3, $s$=1 ($c_i$ = 512, $c_o$ = 1) & [T/32 x 1] \\
& &  & Average Pooling & 1 \\

\bottomrule

\end{tabular}
\end{table*}

\end{document}